*Article*

# Geometric Feature-Based Facial Expression Recognition in Image Sequences Using Multi-Class AdaBoost and Support Vector Machines

**Deepak Ghimire and Joonwhoan Lee \***

Division of Computer Engineering, Chonbuk National University, Jeonju-si, Jeollabuk-do 561-756, Korea; E-Mail: deep@jbnu.ac.kr

**\*** Author to whom correspondence should be addressed; E-Mail: chlee@jbnu.ac.kr; Tel.: +82-63-270-2406; Fax: +82-63-270-2394.

**Abstract:** Facial expressions are widely used in the behavioral interpretation of emotions, cognitive science, and social interactions. In this paper, we present a novel method for fully automatic facial expression recognition in facial image sequences. As the facial expression evolves over time facial landmarks are automatically tracked in consecutive video frames, using displacements based on elastic bunch graph matching displacement estimation. Feature vectors from individual landmarks, as well as pairs of landmarks tracking results are extracted, and normalized, with respect to the first frame in the sequence. The prototypical expression sequence for each class of facial expression is formed, by taking the median of the landmark tracking results from the training facial expression sequences. Multi-class AdaBoost with dynamic time warping similarity distance between the feature vector of input facial expression and prototypical facial expression, is used as a weak classifier to select the subset of discriminative feature vectors. Finally, two methods for facial expression recognition are presented, either by using multi-class AdaBoost with dynamic time warping, or by using support vector machine on the boosted feature vectors. The results on the Cohn-Kanade (CK+) facial expression database show a recognition accuracy of 95.17% and 97.35% using multi-class AdaBoost and support vector machines, respectively.

## 1. Introduction

Automatic facial expression recognition and analysis has been an active topic in the scientific community for over two decades (refer to [1] for a recent review). Recent psychological research has shown that facial expressions are the most expressive way in which humans display emotion. The verbal part of the message contributes only 7% of the effect of the message as a whole, and the vocal part 38%, while facial expression contributes 55% of the effect of the speaker's message [2]. Therefore, automated and real-time facial expression recognition would be useful in many applications, e.g., human-computer interfaces, virtual reality, video-conferencing, customer satisfaction studies, *etc.* in order to achieve the desired result. Although humans detect and interpret faces and facial expressions in a scene with little or no effort, accurate facial expression recognition by machine is still a challenge. Several research efforts have been made regarding facial expression recognition. In general, facial expressions are divided by psychologists into six basic categories: anger, disgust, fear, happiness, sadness, and surprise [3].

The first survey on the research made regarding facial expression recognition was published in 1992 [4], and has been followed up by several researchers [5,6]. A meta-review of the facial expression recognition and analysis challenge has recently been presented in [1], by focusing on clarifying how far the field has come, identifying new goals, and providing the results of the baseline algorithms. It also discusses the future of the field of facial expression recognition, and possible future challenges. Facial expression recognition approaches can be divided into two main categories, based on the type of features used: either appearance-based features, or geometry-based features. Geometry-based features describe the shape of the face and its components, such as the mouth or the eyebrow, whereas appearance-based features describe the texture of the face, caused by expression.

The appearance features that have been successfully employed for emotion recognition are local binary pattern (LBP) operator [7–11], histogram of orientation gradients (HOG) [12,13], local Gabor binary patterns (LGBP) [10], local directional pattern (LDP) [14], non-negative matrix factorization (NMF) based texture feature [15,16], Gabor filter based texture information [16], principle component analysis (PCA) [17], linear discriminant analysis (LDA) [18], *etc.* Among the appearance-based techniques, the theory of NMF has recently led to a number of promising works. In [16], an analysis of the effect of partial occlusion on facial expression recognition is performed, using a method based on Gabor wavelets texture information extraction, supervised image decomposition method based on discriminant NMF (DNMF), and shape-based method. A technique called graph-preserving sparse NMF (GSNMF) was introduced by Zhi *et al.* [15]. The GSNMF is an occlusion-robust dimensionality reduction technique, which transforms high-dimensionality facial expression images into a locality-preserving subspace, with sparse representation. LBP operator is also widely employed by many researchers for the analysis and recognition of facial expression. A comprehensive study on LBP operator-based facial expression recognition was presented by Sahn *et al.* [8]. In this study, they obtained the best facial expression recognition accuracy, by using support vector machine (SVM) classifiers with boosted-LBP features. An extension of the LBP operator, volume LBP (VLBP) and LBP on three orthogonal planes (LBP-TOP) are used in [7] for the recognition of facial expressions. Another example of a system that uses appearance feature to detect facial expressions automatically from video is that by Littlewort *et al.* [19]. Different machine learning techniques are evaluated, and the best result was obtained by selecting a subset of Gabor filters using AdaBoost, and then training SVM on the output of the filters selected by the AdaBoost.

In the geometric feature-based approach the primary step is to localize and track a dense set of facial points. Most geometric feature-based approaches use the active appearance model (AAM) or its variations, to track a dense set of facial points. The locations of these facial landmarks are then used in different ways to extract the shape of facial features, and movement of facial features, as the expression evolves. Choi *et al.* [20] use AAM with second order minimization, and a multilayer perceptron, for the recognition of facial expressions. A recent example of an AAM-based technique for facial expression recognition is presented in [21], in which different AAM fitting algorithms are compared and evaluated. Another example of a system that uses geometric features to detect facial expressions is that by Kotisa *et al.* [22]. A Kanade-Lucas-Tomasi (KLT) tracker was used, after manually locating a number of facial points. The geometric displacement of certain selected candid nodes, defined as the differences of the node coordinates between the first and the greatest facial expression intensity frames, were used as an input to the novel multiclass SVM classifier. Sobe *et al.* [23] also uses geometric features to detect emotions. In this method, the facial feature points were manually located, and Piecewise Bezier volume deformation tracking was used to track those manually placed facial landmarks. They experimented with a large number of machine learning techniques, and the best result was obtained with a simple *k*-nearest neighbor technique. Sung and Kim [24] introduce Stereo AAM (STAAM), which improves the fitting and tracking of standard AAMs, by using multiple cameras to model the 3-D shape and rigid motion parameters. A layered generalized discriminant analysis classifier is then used to combine 3-D shape and registered 2-D appearance, for the recognition of facial expressions. In [25], a geometric features-based approach for modeling, tracking, and recognizing facial expressions on a low-dimensional expression manifold was presented. Sandbach *et al.* [26] recently presented a method that exploits 3D motion-based features between frames of 3D facial geometry sequences, for dynamic facial expression recognition. Onset and offset segment features of the expression are extracted, by using feature selection methods. These features are then used to train the GentleBoost classifiers, and build a Hidden Markov Model (HMM), in order to model the full temporal dynamic of the expressions. Rudovic and Pantic [27] introduce a method for head-pose invariant facial expression recognition that is based on a set of characteristic facial points, extracted using AAMs. A coupled scale Gaussian process regression (CSGPR) model is used for head-pose normalization.

Researchers have also developed systems for facial expression recognition, by utilizing the advantages of both geometric-based and appearance-based features. Lyons and Akamatsu [28] introduced a system for coding facial expressions with Gabor wavelets. The facial expression images were coded using a multi-orientation, multi-resolution set of Gabor filters at some fixed geometric positions of the facial landmarks. The similarity space derived from this representation was compared with one derived from semantic ratings of the images by human observers. In [29], a comparison between geometric-based and Gabor-wavelets-based facial expression recognition using multi-layer perceptorn was presented. Huang *et al.* [30] presented a boosted component-based facial expression recognition method by utilizing the spatiotemporal features extracted from dynamic image sequences, where the spatiotemporal features were extracted from facial areas centered at 38 detected fiducial interest points. The facial points were detected by using ASM, and the features at those points were described by using LBP-TOP operator.

In the current paper, two methods are proposed for recognizing dynamic facial expressions, either directly by using multiclass AdaBoost, or by using SVM on the boosted geometric features. Let us consider a video shot containing a face, whose facial expression evolves from a neutral state to a fully expressed state. The facial expression recognition is performed using only geometric information extracted from the sequence of facial images, without taking into consideration any facial texture information. The proposed facial expression recognition system is fully automatic, in which elastic bunch graph (EBG) is used to initialize facial landmarks at the first frame. Multi-resolution, multi-orientation Gabor filter responses at each facial landmark are now used to track the landmarks in the consecutive frames. The result of landmarks tracking is now normalized in such a way that the landmark positions on the first frame for each facial expression are the same, and follows the movement of the landmarks as the expression evolves over time, from neutral state to its highest intensity. A feature vector corresponding to each landmark point, as well as feature vectors from each pair of landmarks, are now created, and considered as a feature pool. We have a large number of feature vectors in the feature pool, but only some of them provide discriminative information for recognizing facial expressions. Therefore, a subset of feature vectors are chosen, by applying the AdaBoost algorithm for feature selection, with the help of dynamic time warping (DTW) similarity distance [31,32] between input feature vector and prototypic feature vector, of each facial expression. The prototype for each facial expression is formed by taking the median of all the tracked landmark positions of the corresponding facial expression dataset. The facial expression can now be recognized, by using multiclass AdaBoost with DTW similarity distance between feature vectors, or by using support vector machine on the subset of features selected by AdaBoost. There is always an issue relating to which features are important for distinguishing each facial expression from the rest of the facial expressions, either in the case of geometric features, or in the case of appearance features. A one-versus-all class classification scheme is used, and the subset of feature vectors from the feature pool is selected for each facial expression, and analyzed in terms of landmarks from different regions of the face, which are used to generate the feature vectors. The experiments were performed on the CK+ facial expression database, and the result shows that the proposed facial expression recognition system can achieve good recognition accuracy up to 97.35% using a small subset of boosted-features, when recognizing six basic facial expressions using the SVM classifier.

This paper is organized as follows: Section 2 describes the landmark initialization process on the neutral frame (first frame of the video shot), tracking those landmarks over time as the expression evolves, and the normalization of tracking results of the landmarks. The proposed geometric features are described in Section 3. Subset of feature vector selection using AdaBoost with DTW similarity as a weak classifier is described in Section 4. The experimental results are presented in Section 5. Finally, Section 6 concludes the proposed facial expression recognition system.

## 2. Landmark Initialization, Tracking and Normalization

Facial expression recognition is mainly composed of three subsystems: facial landmark tracking, building features from the landmark tracking result, and classification of the extracted features.

*2.1. Landmark Initialization and Tracking Using Elastic Bunch Graph*

The elastic graph matching (EGM) method was first proposed by Lades *et al.* [33], and applied to face recognition. Based on the work of Lades *et al.*, Wiskott *et al.* [34] extracted more than one feature on one landmark point, called it EBG, and applied it to face recognition. Our implementation of EBG based landmark initialization is based on the algorithm developed by Wiskott *et al.* which was included by Colorado State University (CSU) as a baseline algorithm for comparison of face recognition algorithms [35]. The algorithm starts by creating a bunch graph. Each node of the bunch graph corresponds to a facial landmark, and contains a bunch of model jets (Gabor jets) extracted from the model imagery. The Gabor jets are a collection of complex Gabor coefficients from the same location in an image. The coefficients are generated using Gabor wavelets, of a variety of different sizes, orientations and frequencies. The bunch graph serves as a database of landmarks descriptors that can be used to locate landmarks in the novel imagery.

Figure 1 shows the overall process of the landmark initialization and tracking. Locating a landmark in a novel imagery has two steps. First, the location of the landmark is estimated based on the known locations of other landmarks in the image, and second, that estimate is refined by extracting a Gabor jet from that image on the approximate locations and comparing that jet to one of the models. To make the system fully automatic the approximate locations of at least one or two landmarks are needed at the beginning. This goal is achieved by first localizing the face region in the image using the Haar-like feature based face detection method proposed in [36]. Now the landmarks at the center of the two eyes are searched within a face region by using the same Haar-like feature based object detection method proposed by Viola and Jones [36]. Estimating the location of the other landmarks is easy, based on the known eye coordinates. Each new landmark location is estimated based on the set of previously localized points. The landmark location is then refined by comparing a Gabor jet extracted from the estimated point to a corresponding model jet from the bunch graph. The process is iterated until all landmark locations are found. 52 landmarks in the neutral face image (first frame of the video shot) are initialized.

**Figure 1.** Overall block diagram of the landmark initialization and tracking process.

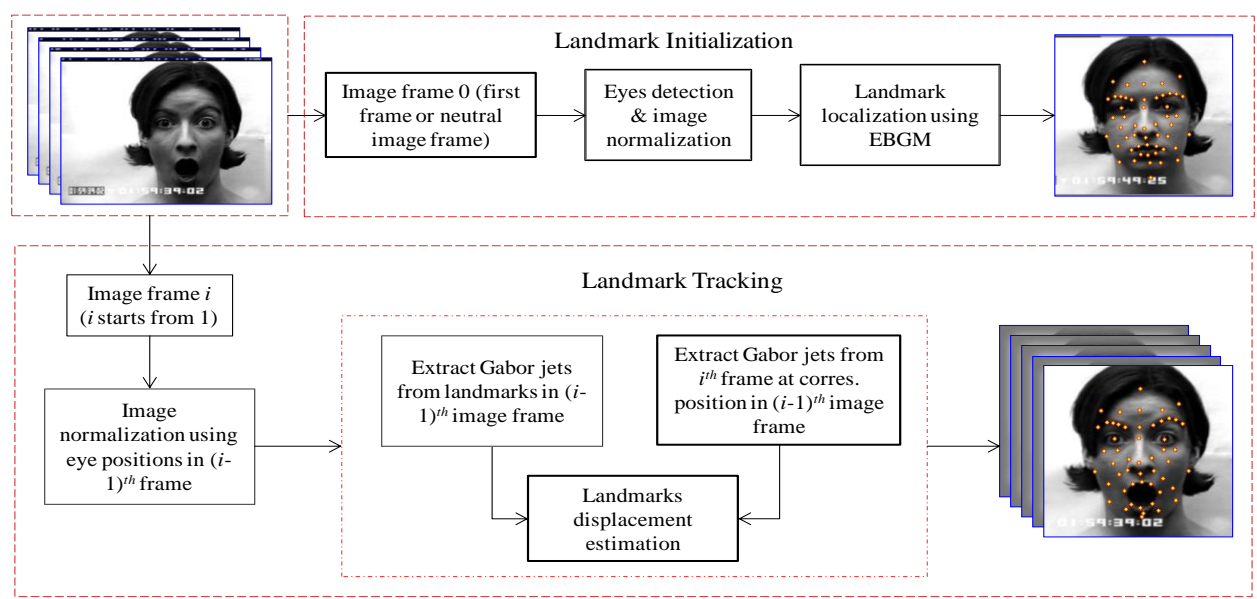

Once the landmarks are initialized in the first frame of the video shot, the next step is to track them over time, as expression evolves. In each incoming frame, the displacement of landmarks with respect to the previous frame should be estimated. The Gabor jet corresponding to each landmark in the neutral frame is already computed. The Gabor jet from the next frame for each landmark, as in the same landmark position of the previous frame, is now extracted. The displacement of the landmarks with respect to the previous frame can be directly calculated from those two Gabor jets using the equation for direct displacement estimation given in [34]. This displacement gives the exact position of the landmark in the current frame. The Gabor jet for each landmark in the current frame is now updated, and the same process is repeated, to find the displacement of the landmarks in the next frame. The promising result of landmark tracking is obtained by using this concept. Figure 2 shows the result of 52 landmarks tracking in some sequences of facial expression; only a few images from each sequence are shown in the figure. The first, second and third row belong to anger, disgust and surprise facial expressions, respectively.

**Figure 2.** Examples of the result of facial landmark tracking.

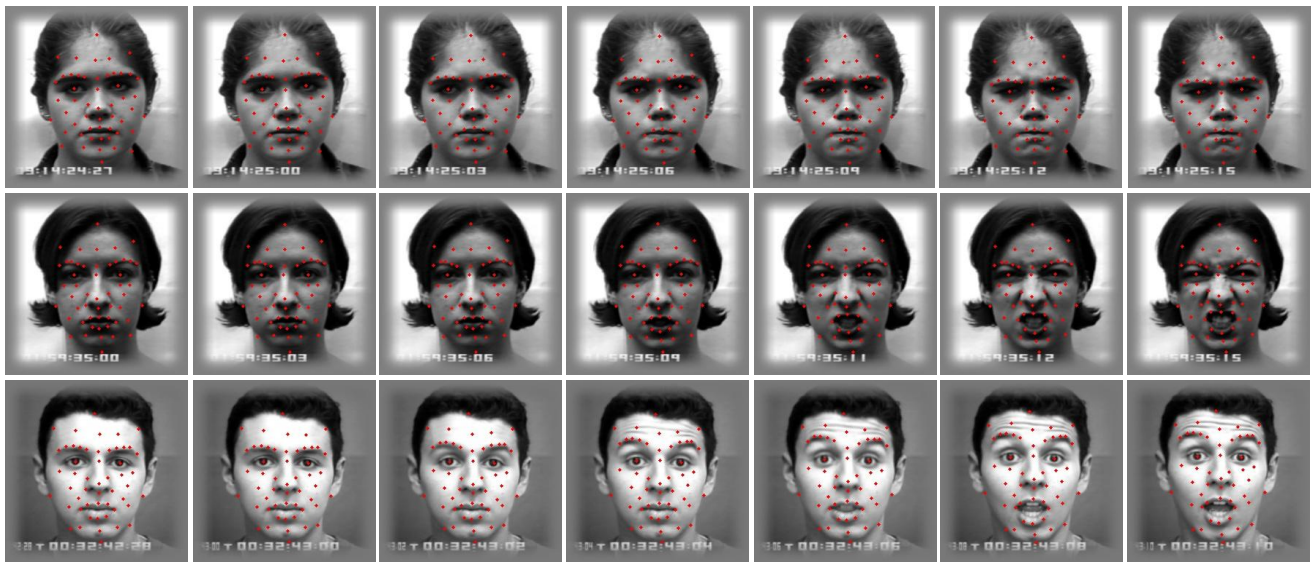

## 2.2. Landmark Normalization

The landmark normalization brings each landmark to the uniform coordinate position in the first frame of the video shot, and as the expression evolves, the landmarks are displaced accordingly. Let us suppose $S_i^k$ is the tracking result of the *i*th landmark in the *k*th expression sequences:

$$S_i^k = \left\{(x_0, y_0)_i^k, (x_1, y_1)_i^k, (x_2, y_2)_i^k, ...., (x_{N-1}, y_{N-1})_i^k, (x_N, y_N)_i^k\right\} \qquad (1)$$

where $(x_l, y_l)_i^k$ is the *i*th landmark coordinate position in the *l*th frame of the *k*th expression sequence, and *N* is the number of frame in an expression sequence.

An average landmark position corresponding to each landmark is computed from all the neutral images, which is the first frame in each video shot. Suppose $(\mu_{x0}, \mu_{y0})_i$ denotes the average landmark position of the *i*th landmark in the first frame of the expression sequences. For each tracking result of the facial expression whose landmarks are to be normalized, the difference between the first frame landmark and the average landmark is determined. This gives the displacement of the landmark, with respect to the average landmark position. Suppose $(\delta_{x0}, \delta_{y0})_i^k$ denotes the displacement of the *i*th landmark in the first frame of the *k*th expression sequence, with respect to the average landmark position:

$$(\delta_{x0}, \delta_{y0})_i^k = (\mu_{x0} - x_0, \mu_{y0} - y_0)_i^k. \quad (2)$$

The displacement corresponding to each landmark is now added to the landmark positions in every frame of the facial expression sequence. The transformed result of landmark tracking is now denoted by $S_i'^k$, and is defined as:

$$S_i'^k = \{(x_0 + \delta_{x0}, y_0 + \delta_{y0})_i^k, (x_1 + \delta_{x0}, y_1 + \delta_{y0})_i^k, ..., (x_N + \delta_{x0}, y_N + \delta_{y0})_i^k\}. \quad (3)$$

The tracking result is normalized in such a way that each landmark for all the expression sequence now starts from the same coordinates position, *i.e.*, $(\mu_{x0}, \mu_{y0})_i$ and evolves according to the displacement in the succeeding frames. Figure 3 shows the tracking result of landmarks before (first row), and after (second row), normalization (note the edges between landmarks are given just to make a face-like appearance).

**Figure 3.** Example of landmark tracking sequences before (**First** row) and after (**Second** row) normalization.

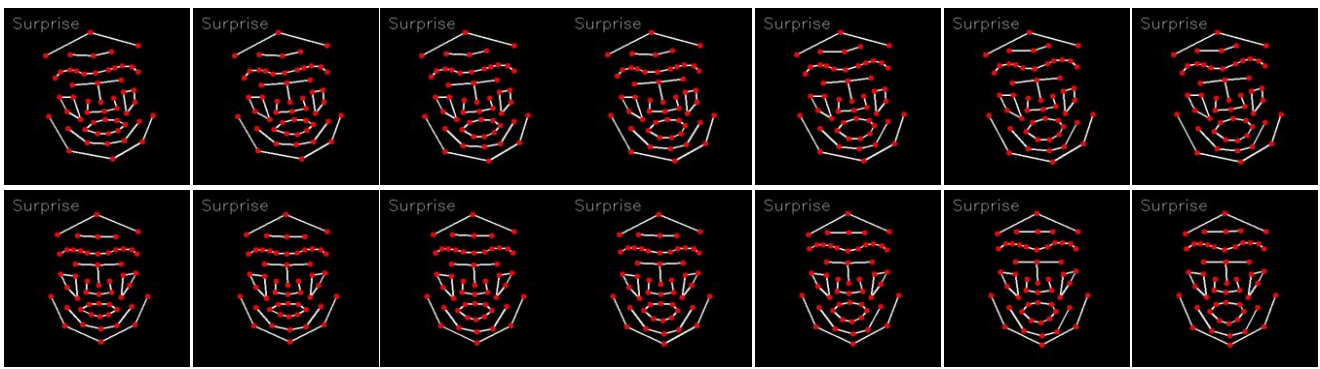

## 3. Feature Extraction

In the proposed approach, the facial expression recognition is performed based only on geometrical information, without directly taking any facial texture information. The feature pool is made from two types of features, one by considering the tracking result of each single facial landmark, and the other by considering the tracking result of pairs of facial landmarks. Suppose $(x', y')$ is the transformed landmark coordinate position, and let us rewrite Equation (3) in the following form:

$$S_i'^k = \{(x_0', y_0')_i^k, (x_1', y_1')_i^k, ..., (x_N', y_N')_i^k\}. \quad (4)$$

The number of frames in different video shots of facial expression can be different. As we will describe in the next section, the subset of feature selection from the available feature pool is performed by using AdaBoost, with DTW similarity between prototypic feature vector and input feature vector as a weak classifier. Therefore, for each facial expression class, a prototypic facial expression should be created. For this purpose, the number of frames should be equal, in each facial expression sequence in the database. Once the prototype creation and feature selection is completed, the numbers of frames in the sequence need not necessarily be equal in the classification stage. Therefore, let us consider in Equation (4) that each facial expression has the same number of frames. In practice, we resize the landmark tracking sequence, using linear interpolation. In our experiment we use $N = 15$, and there are $L = 52$ facial landmarks. Feature type one is a feature vector from the individual landmark tracking sequence. Each landmark coordinate in a sequence is now subtracted from the first landmark coordinate, *i.e.*, landmark position in the neutral frame, to create the type one feature vector. Suppose $(\delta x'_l, \delta y'_l)^k_i$ denotes the difference in the *i*th landmark in the *l*th frame, from the *i*th landmark in the first frame of the *k*th video shot:

$$(\delta x'_l, \delta y'_l)^k_i = (x'_l - x'_0, y'_l - y'_0)^k_i. \tag{5}$$

Equation (6) defines the type one feature vector:

$$\delta S'^k_i = \{(\delta x'_1, \delta y'_1)^k_i, (\delta x'_2, \delta y'_2)^k_i, ..., (\delta x'_N, \delta y'_N)^k_i\}. \tag{6}$$

Next, the feature vector is created from the pair of landmarks in the expression sequence. We call this feature a type two feature vector. First, the angle and Euclidian distance between each pair of landmarks within a frame are calculated. Suppose $(d'_l, \theta'_l)^k_{i,j}$ denotes the distance and angle between the *i*th and *j*th pair of landmarks in the *l*th frame of the *k*th expression sequence. Let us denote the calculated sequence of distance and angle as $G'^k_{i,j}$.

$$G'^k_{i,j} = \{(d'_0, \theta'_0)^k_{i,j}, (d'_1, \theta'_1)^k_{i,j}, ..., (d'_N, \theta'_N)^k_{i,j}\}. \tag{7}$$

Now, those distance and angles are subtracted from the corresponding distance and angles on the first frame of the video shot. Suppose $(\delta d'_l, \delta \theta'_l)^k_{i,j}$ denotes the change in distance and angle between the *i*th and *j*th pair of landmarks in the *l*th frame, with respect to the first frame, in the *k*th video shot:

$$(\delta d'_l, \delta \theta'_l)^k_{i,j} = (d'_l - d'_0, \theta'_l - \theta'_0)^k_{i,j}. \tag{8}$$

Equation (9) defines the type two feature vector:

$$\delta G'^k_{i,j} = \{(\delta d'_1, \delta \theta'_1)^k_{i,j}, (\delta d'_2, \delta \theta'_2)^k_{i,j}, ..., (\delta d'_N, \delta \theta'_N)^k_{i,j}\}. \tag{9}$$

A type one feature has $L = 52$ feature vectors, and type two feature has $M = L \times (L-1)/2 = 1,326$ feature vectors. In total, there are $L + M = 1,378$ feature vectors in the feature pool. Out of those feature vectors, only some of them carry most of the discriminative information for the recognition of facial expressions. The AdaBoost feature selection scheme is used to select the subset of feature vectors from the feature pool that is sufficient for the recognition of the facial expressions.

## 4. Feature Selection Using AdaBoost

The AdaBoost learning algorithm proposed by Freund and Schapire [37], in its original form, is used to boost the classification performance of a simple learning algorithm. It does this by combining a collection of weak classification functions, to form a stronger classifier. In the language of boosting, the simple learning algorithm is called a weak classifier. AdaBoost is not only a fast classifier; it is also a feature selection technique. An advantage of feature selection by AdaBoost, is that features are selected contingent on the features that have already been selected. Different types of appearance based feature selection using AdaBoost for facial expression recognition can be found in the literature [11,19,38,39]. In our system, a variant of multi-class AdaBoost proposed by Jhu *et al.* [40] is used, to select the important features sufficient for facial expression recognition from the feature pool defined in Section 3. In our system, classifier learning is not necessary, because the weak classifier is based on the DTW similarity [31,32] between the prototypical feature vector, and the input feature vector. Before describing the feature selection using AdaBoost, let us first describe the creation of prototypical feature vector, and DTW similarity, which will serve as a weak classifier.

Prototypical feature vectors for each class of facial expression are created by taking the median of each corresponding elements in the feature vectors, from all the training set of those facial expression sequences. Here the assumption is that each class of facial expression can be modeled by using unimodal distribution, and this is the proper assumption which is supported by the experimental results. The median is selected, instead of the mean, because it is less affected by the presence of an outlier. Let $U$ be the facial expression database that contains facial expression sequences. The database is clustered into six different classes, $U_c$, $c = 1,…,6$, with each representing one of six basic facial expressions (anger, disgust, fear, happiness, sadness, and surprise). Suppose the prototypical feature vector of class $c$ for type one feature vectors and type two feature vectors is denoted by $P_{U_c}(\delta S'_i)$ and $P_{U_c}(\delta G'_{i,j})$, respectively:

$$P_{U_c}(\delta S'_i) = \underset{k \in U_c}{median} \left\{ (\delta x'_1, \delta y'_1)^k_i, (\delta x'_2, \delta y'_2)^k_i, ..., (\delta x'_N, \delta y'_N)^k_i \right\} \quad (10)$$

$$P_{U_c}(\delta G'_{i,j}) = \underset{k \in U_c}{median} \left\{ (\delta d_1, \delta \theta_1)^k_{i,j}, (\delta d_2, \delta \theta_2)^k_{i,j}, ..., (\delta d_N, \delta \theta_N)^k_{i,j} \right\} \quad (11)$$

Figure 4 shows the maximum intensity frame of each class of prototypical facial expression sequences.

**Figure 4.** Maximum intensity frame of each facial expression prototype.

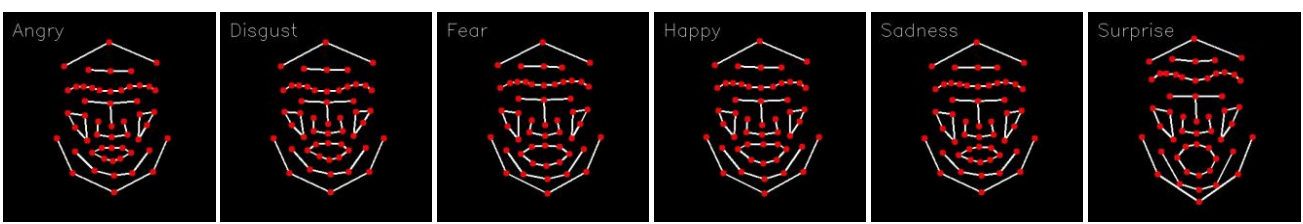

Our goal is to search a small number of the most discriminative feature vectors from the feature pool. In support of this goal, the weak classifier is designed to select a single feature vector, which best classifies the training data. The class level of the feature vector is decided based on the minimum DTW similarity distance, with prototypical feature vectors. The DTW is a well-known algorithm that aims to compare and align two sequences of data points. It is worth noting that the sequences can be of different length. The sequences are "warped" non-linearly in the time dimension, to determine a measure of their similarity, independent of certain non-linear variations in the time dimensions. Distances like Euclidian, Manhattan, *etc.* that align the *i*th point on one time series with the *i*th point on the other, will produce a poor similarity score, whereas non-linear alignment produces a more intuitive similarity measure, allowing similar shapes to match, even if they are out of phase on the time axis. Although DTW was originally developed for speech recognition [31], it has also been applied to many other fields. In our system, the DTW is an efficient method for finding similarity between feature vectors, because the length of the feature vector can be different, according to the number of frames in the facial expression sequence, as well as in different persons, as the movement of landmark positions is non-linear as the facial expression evolves. We use the DTW algorithm used in [32] to quickly find the similarity between two sequences. Now, a weak classifier ($T(x, x^{(p)}, f)$) consists of a feature vector (*f*), input facial expression (*x*) and prototypical facial expressions ($x^{(p)}$):

$$T(x, x^{(p)}, f) = \arg\min_c \{d_{DTW}(f(x), f(x_c^{(p)}))\}. \tag{12}$$

In practice, no single feature can perform this classification task with low error. Features that are selected in the early process yield less classification error rates, than features selected in the later rounds. Algorithm 1 shows the variant of multi-class AdaBoost learning algorithm proposed in [40], which they term Stagewise Additive Modeling using a Multi-class Exponential (SAMME) loss function.

The multi-class AdaBoost algorithm given in Algorithm 1 is very similar to AdaBoost, with the major difference in Equation (13). Now, in order for $\alpha^{(m)}$ to be positive, we only need $(1 - err^{(m)}) > 1/K$, or the accuracy of each weak classifier to be better than random guessing, rather than $1/2$. The significance of extra term $\log(K-1)$ in Equation (13) is explained in [40].

**Algorithm 1.** Multi-class AdaBoost learning algorithm. *M* hypothesis are constructed each using a single feature vector. The final hypothesis is a weighted linear combination of *M* hypothesis.

1. *Initialize the observation weights $w_{1,i} = 1/n$, $i = 1, 2, ..., n$.*
2. *For m = 1 to M:*
   a. *Normalize the weights, $w_{m,i} \leftarrow w_{m,i} / \sum_{j=1}^{n} w_{m,j}$.*
   b. *Select the best week classifier with respect to the weighted error*
   $$err^{(m)} = \min_f \sum_{i=1}^{n} w_i \cdot I(c_i \neq T(x_i, x^{(p)}, f)) / \sum_{i=1}^{n} w_i.$$
   c. *Define $T^{(m)}(x) = T(x, x^{(p)}, f_m)$ where $f_m$ is the minimize of $err^{(m)}$.*
   d. *Compute*
   $$\alpha^{(m)} = \log \frac{1 - err^{(m)}}{err^{(m)}} + \log(K - 1). \tag{13}$$
   e. *Update the weights: $w_{m,i} \leftarrow w_{m,i} \cdot \exp(\alpha^{(m)} \cdot I(c_i \neq T^{(m)}(x_i)))$, $i = 1, ..., n$.*
3. *The final strong classifier is:*
   $$C(x) = \arg\max_k \sum_{m=1}^{M} \alpha^{(m)} \cdot I(T^{(m)}(x) = k). \tag{14}$$

Figure 5 shows the first few features selected by multi-class AdaBoost. The blue dot indicates a type one feature vector generated from that landmark, and each line connecting a pair of landmarks indicates a type two feature vector generated from that pair of landmarks. Most of the selected feature vectors belong to type two features. This proves that the movements of landmarks, as the expression evolves for each facial expression, are not independent.

**Figure 5.** The first few (20, 40, 60, 80, and 100) features selected using multi-class AdaBoost. A blue dot indicates a feature vector extracted only from that landmark tracking result, as the expression evolves over time; whereas a red line connecting two landmarks indicates the feature vector extracted from that pair of landmarks tracking result, as the expression evolves over time.

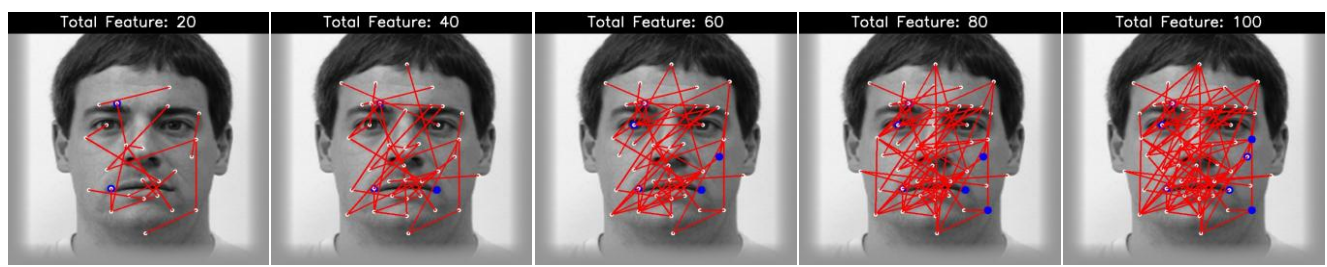

Shan *et al.* [8] applied the AdaBoost algorithm to determine a subset of the LBP histogram for each facial expression. As each LBP histogram is calculated from a sub-region, AdaBoost is actually used to find the sub-regions that contain more discriminative information for facial expression, in terms of the LBP histogram. Here, we are also interested to find the subset of feature vector for each facial expression. This will give the information about what are the features that are important to discriminate one particular class of facial expression, from the rest of the facial expression classes. It is worth noting that when $K = 2$, the feature selective multi-class AdaBoost algorithm in Algorithm 1 reduces to a two-class AdaBoost algorithm. As our weak classifier is based upon the DTW similarity of the feature vector with the prototypic feature vector, in the two-class case, we can have a prototypical facial expression for the positive class, but it is not feasible to make a single unimodel prototype for the negative class of facial expressions. For example, if we want to find the features for the angry facial expression, the anger is the positive class, whereas the rest of facial expressions (disgust, fear, happiness, sadness and surprise) belong to the negative class. Therefore, we keep all the prototypical facial expressions as they are, and use a single prototype for the positive class of facial expressions, whereas we use multiple prototypes (one for each class of facial expression) for the negative class of facial expressions. The weak classifier now classifies input facial expression into the positive class, if the DTW similarity distance is minimum with the positive class of prototype among all the prototypes; otherwise it classifies the input facial expression into the negative class. Figure 6 shows the first few features selected for each class of facial expressions, using this scheme. It is observed that different classes of facial expression have different key discriminative geometric features.

In general, for anger, disgust and sadness facial expressions, usually the mouth is closed, and the landmarks movement is smaller, as compared with the movements of the landmarks in the case of fear, happiness and surprise facial expressions. From Figure 6, it seems that the selected feature vector for anger, sadness and disgust are from the near pair of landmarks, whereas in the case of fear, happiness and surprise, most of the feature vectors are from the far pair of landmarks. Actually, it is hard to analyze the landmarks, to determine which region of the face carries the discriminative information for each class of facial expressions, by just looking at the selected features. Therefore, we divided the landmarks into different subsets, according to the specific region, in order to determine regions or pairs of regions that contain discriminative landmarks for each facial expression. Figure 7 shows the grouping of landmarks, according to the face region, into seven subsets.

**Figure 6.** The first few (20, 40, and 60) features selected by AdaBoost for each class of facial expression. A red dot indicates a feature vector extracted from the tracking result of that landmark, as the facial expression evolves over time, whereas a red line connecting two landmarks indicates a feature vector extracted from the tracking result of those pair of landmarks, as the facial expression evolves over time.

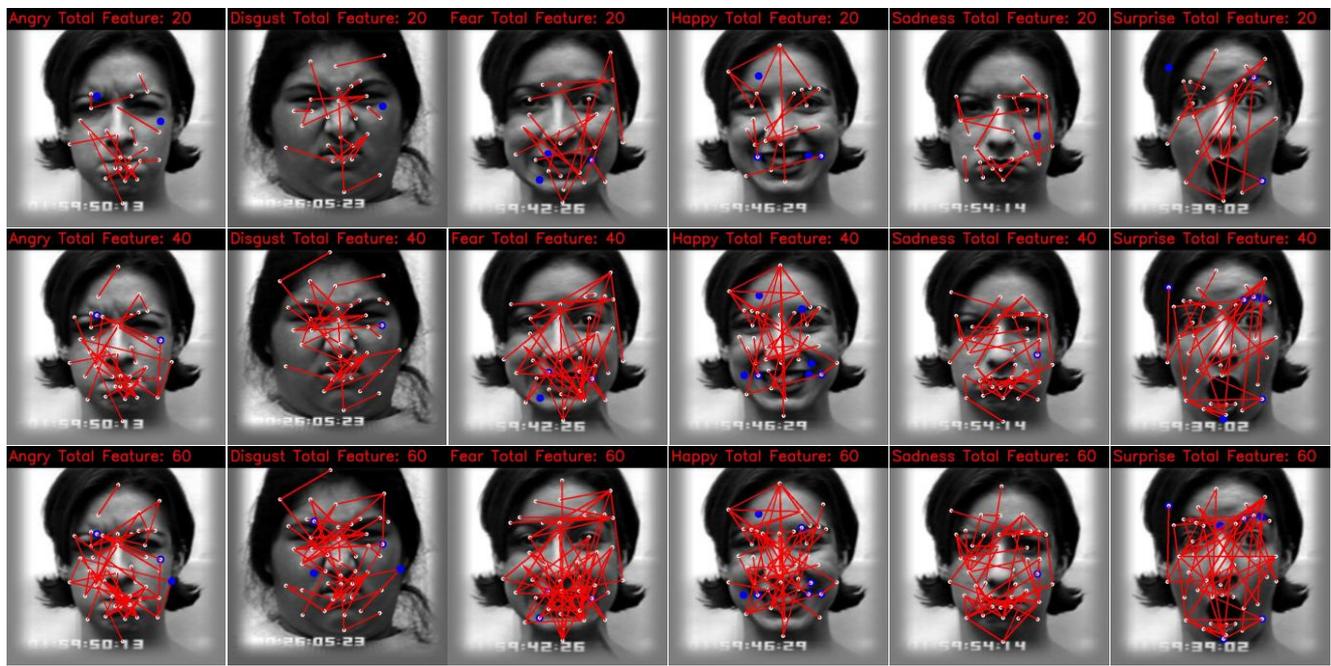

**Figure 7.** Grouping of landmarks into different regions according to the facial geometry.

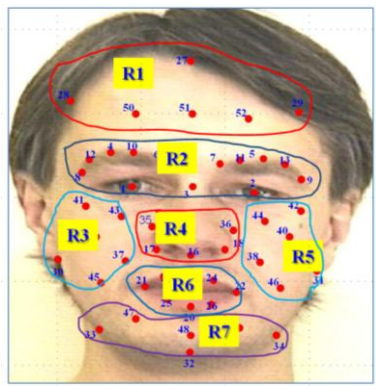

The selected features for each facial expression are either from the pair of landmarks within a region, or from the pair of landmarks from two regions. Table 1 lists the region, as well as the pair of regions, from which the landmarks are chosen to extract the most discriminative feature vector for each facial expression. A single or pair of landmarks within region R2, which is the eye and eyebrow region, is used to build the discriminative feature vectors for all the facial expressions. Pairs of landmarks from region R1–R2, and R2–R3 also have the discriminative information for almost all facial expressions. Region R6, the mouth region, contains discriminative landmarks for anger and sadness facial expressions. Table 1 shows more detail of the different regions of the face, from which the landmarks tracking results are used to build most of the discriminative feature vectors.

**Table 1.** Different regions (single/pair) of the face, with the most discriminative facial landmarks. Single region (Ri) means landmarks within this region are used to build a feature vector either from the stand alone landmark tracking result or pair of landmark tracking result. Pair of region (Ri–Rj) means two landmarks, one from each, is used to build the feature vector from the tracking result of those landmarks.

| Expression | Face Regions (Single/Pair) with the Most Discriminative Landmarks |
|---|---|
| Anger | R1-R2, R2, R2-R3, R5-R7, R6, R6-R7 |
| Disgust | R1-R2, R2, R2-R3, R5, R5-R6, R5-R7 |
| Fear | R1-R2, R2, R3-R6, R5-R6, R2-R7, R4-R7, R6-R7 |
| Happiness | R1-R2, R2, R2-R4, R2-R6, R3-R6, R4-R6 |
| Sadness | R2-R3, R2, R3-R7, R5-R6, R6, R5-R7, R6-R7 |
| Surprise | R1-R3, R2, R2-R3, R2-R7, R2-R5, R7 |

## 5. Experimental Results

*5.1. Dataset Description*

The Extended Cohn-Kanade (CK+) database [41] was used for facial expression recognition in six basic facial expression classes (anger, disgust, fear, happiness, sadness, and surprise). This database consists of 593 sequences from 123 subjects. The image sequence varies in duration (*i.e.*, seven to 60 frames), and incorporates the onset (which is also the neutral face) to peak formation of the facial expression. Image sequences from neutral to target display were digitized into 640 × 480 or 640 × 490 pixel arrays. Only 327 of the 593 sequences have a given emotional class. This is because these are the only ones that fit the prototypic definition. For our study, 315 sequences of the data set are selected from the database, for basic facial expression recognition.

The most usual approach for testing the generalization performance of a classifier is the K-fold cross validation approach. A five-fold cross validation was used in order to make maximum use of the available data, and produce averaged classification accuracy results. The dataset from each class of facial expression was divided into five subsets for the entire process, even for obtaining the prototypical facial expressions. Each time, one of the five subsets from each class was used as the test set, and the other four subsets were put together to form a training set. The classification accuracy is the average accuracy across all five trials. To get a better picture of the recognition accuracy of the individual expression type, the confusion matrices are given. The confusion matrix is an $n \times n$ matrix, in which each column of the matrix represents the instances in a predicted class, while each row represents the instances in an actual class. The diagonal entries of the confusion matrix are the rates of facial expressions that are correctly classified, while the off-diagonal entries correspond to misclassification rates.

*5.2. Facial Expression Recognition Using Multi-Class AdaBoost*

In our system, the variant of multi-class AdaBoost proposed in [40] is used to select the discriminative feature vectors extracted from landmark tracking results. At the same time, the AdaBoost algorithm determines the weights associated with each feature vector. The facial expressions are recognized by using the strong classifier given in Equation (14). The classification is based on the DTW similarity distance between the selected feature vectors of the test facial expression sequence, with the feature vector associated with each class of prototypic facial expression sequence. Each feature vector classifies the facial expression into one of the six classes, according to the minimum DTW similarity distance. One of the advantages of using DTW similarity measurement is that the two feature vectors to be compared need not necessarily be of equal length. Increasing the number of features in Equation (14) also increases the classification accuracy up to some limit. Figure 8 shows a graph of the number of features *versus* recognition accuracy, for both training and testing data.

**Figure 8.** Recognition accuracy under different numbers of AdaBoost selected features.

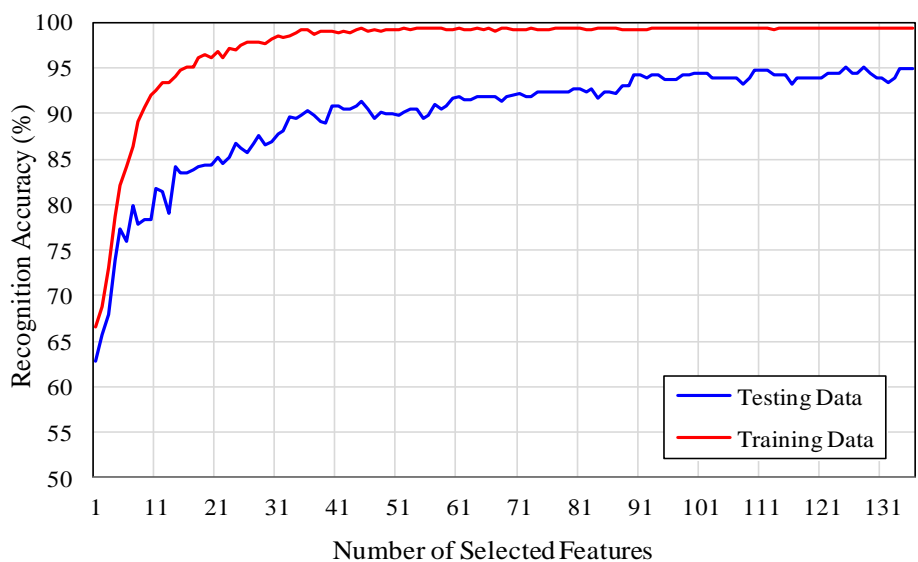

**Table 2.** Confusion matrix for facial expression recognition in percentages, using multi-class AdaBoost with 75 feature vectors.

|           | **Anger** | **Disgust** | **Fear** | **Happiness** | **Sadness** | **Surprise** |
|-----------|-----------|-------------|----------|---------------|-------------|--------------|
| Anger     | 90        | 5           | 0        | 0             | 5           | 0            |
| Disgust   | 0         | 95          | 0        | 5             | 0           | 0            |
| Fear      | 0         | 0           | 84       | 12            | 0           | 4            |
| Happiness | 0         | 0           | 3.08     | 96.92         | 0           | 0            |
| Sadness   | 6.67      | 0           | 3.33     | 0             | 90          | 0            |
| Surprise  | 0         | 0           | 0        | 1.25          | 0           | 98.75        |

With 52 landmark tracking results there are a total of 1,378 possible feature vectors, but only a few of them are sufficient to discriminate the six basic facial expressions. The highest classification accuracy of 95.17% is achieved with a minimum of 125 feature vectors. Tables 2, 3 show the confusion matrix of facial expression recognition, using multi-class AdaBoost with 75 and 125 number of feature vectors, respectively. Some of the fear and happiness facial expression are confused with each other. The discrimination of happiness and fear failed, because these expressions had a similar motion of mouth. The sadness and anger facial expressions are even difficult to recognize by a human observer. Some of the disgust facial expression was misclassified as happiness.

**Table 3.** Confusion matrix for facial expression recognition in percentages, using multi-class AdaBoost with 125 feature vectors.

|  | **Anger** | **Disgust** | **Fear** | **Happiness** | **Sadness** | **Surprise** |
|---|---|---|---|---|---|---|
| Anger | 95 | 5 | 0 | 0 | 0 | 0 |
| Disgust | 0 | 95 | 1.67 | 3.33 | 0 | 0 |
| Fear | 0 | 0 | 92 | 8 | 0 | 0 |
| Happiness | 0 | 0 | 3.08 | 96.92 | 0 | 0 |
| Sadness | 6.67 | 0 | 0 | 0 | 93.33 | 0 |
| Surprise | 0 | 0 | 0 | 1.25 | 0 | 98.75 |

**Figure 9.** Average confusion score of week classifiers in percentages, for the recognition of each class of facial expressions.

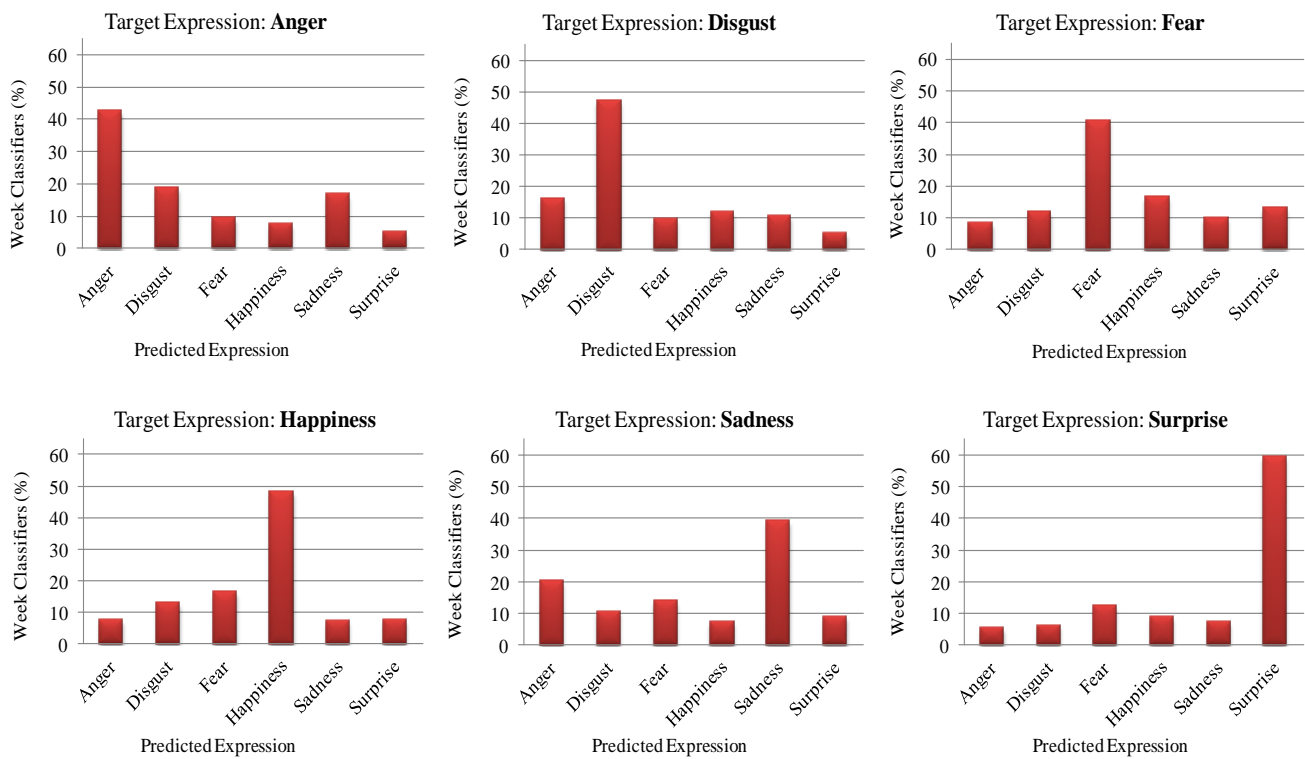

None of the individual feature vectors can classify facial expressions with high accuracy. The strong classifier (Equation (14)) classifies any input expression to one of six basic classes, which has the highest confidence score. It is interesting to know for each facial expression what percentage of the feature vectors is classified correctly, and what percentage of the feature vectors is misclassified. This will give more precise information about the confusion score for each facial expression. Figure 9 shows the confusion scores of weak classifiers in percentages, for each class of facial expression. For example, it can be seen from the figure that 42.52% of the feature vectors classify angry facial expressions correctly, whereas 18.74% of the features vector classify them as disgusted, 9.44% of the feature vector classify them as fearful, 7.61% of the feature vectors classify them as happy, 16.70% of the feature vectors classify them as sad, and finally 4.98% of the feature vectors classify them as surprised facial expressions. In case of the surprised facial expressions, the highest percentage, *i.e.*, 59.51% of feature vectors classify them correctly. Therefore, we can see that, mostly, angry facial expressions are confused with disgusted and sad; disgusted facial expressions are confused with angry; fearful facial expressions are confused with happy and surprised, happy facial expressions are confused with fearful and disgusted; sad facial expressions are confused with angry; and surprised facial expression are confused with fearful.

*5.3. Facial Expression Recognition Using SVM with Boosted Features*

The SVM is a class of linear classification algorithm, which aims to find a separating hyperplane, with as wide a margin as possible, between two different categories of data. In our experiment, we use a publically available implementation of SVM, *libsvm* [42], at which we used radial basic function (RBF) kernel and the optimal parameter selection is done based on the grid search strategy [43].

As given in Equations (6) and (9) in Section 3, these are the two types of features from the landmark tracking results used in our system. Once the AdaBoost selects the feature vectors, for the classification of facial expressions using SVM, we generate the new set of features from the set of feature vectors selected using AdaBoost. In order to keep the dimensionality of the feature used in SVM classification as small as possible, we only took the maximum displacement value in two directions from the feature vector defined in Equation (6), and the maximum change in angle and distance from the feature vector defined in Equation (9). The feature vector defined in Equation (6), which is associated with the *i*th landmark tracking result of the *k*th facial expression sequence, gives the following two values:

$$\delta x'^k_{i\ max} = \max\{\delta x'^k_{1,i}, \delta x'^k_{2,i},..., \delta x'^k_{N,i}\}$$
$$\delta y'^k_{i\ max} = \max\{\delta y'^k_{1,i}, \delta y'^k_{2,i},..., \delta y'^k_{N,i}\}.$$

Similarly, the feature vector defined in Equation (9), which is associated with the *i*th and *j*th landmark tracking result of the *k*th facial expression sequence, gives the following two values:

$$\delta d'^k_{(i,j)max} = \max\{\delta d'^k_{1(i,j)}, \delta d'^k_{2(i,j)},..., \delta d'^k_{N(i,j)}\}$$
$$\delta \theta'^k_{(i,j)max} = \max\{\delta \theta'^k_{1(i,j)}, \delta \theta'^k_{2(i,j)},..., \delta \theta'^k_{N(i,j)}\}.$$

The output of this process is a single feature vector for each video. The dimensionality of the feature depends upon the number of features selected by AdaBoost. If we use *L* feature vectors selected by AdaBoost, the dimensionality of feature for SVM classification will be $L \times 2$. The experiments show that more than 90% features have been selected from type two features, *i.e.*, feature vectors generated by a pair of landmarks tracking results. This proves that the movements of landmarks, as the particular facial expression evolves, are not independent. Tables 4–6 show the confusion matrices for facial expression recognition using 100, 200 and 400 AdaBoost selected features, with dimensionality 200, 400, and 800 respectively. The average recognition accuracy was 93.20%, 95.50% and 97.35%, using 200, 400 and 800 dimensional features, respectively. There is an improvement of 2.18% for facial expression recognition using SVM with boosted features, over facial expression recognition using AdaBoost with DTW similarity distance.

**Table 4.** Confusion matrix for facial expression recognition in percentages, using SVM with boosted features (100 AdaBoost selected features).

|  | **Anger** | **Disgust** | **Fear** | **Happiness** | **Sadness** | **Surprise** |
|---|---|---|---|---|---|---|
| Anger | 92.5 | 2.5 | 0 | 0 | 5 | 0 |
| Disgust | 0 | 96.67 | 1.67 | 0 | 1.67 | 0 |
| Fear | 0 | 0 | 92 | 8 | 0 | 0 |
| Happiness | 0 | 3.08 | 1.54 | 95.38 | 0 | 0 |
| Sadness | 6.67 | 0 | 3.33 | 0 | 87.67 | 3.33 |
| Surprise | 0 | 0 | 2.5 | 0 | 2.5 | 95 |

**Table 5.** Confusion matrix for facial expression recognition in percentages, using SVM with boosted features (200 AdaBoost selected features).

|  | **Anger** | **Disgust** | **Fear** | **Happiness** | **Sadness** | **Surprise** |
|---|---|---|---|---|---|---|
| Anger | 100 | 0 | 0 | 0 | 0 | 0 |
| Disgust | 1.67 | 96.67 | 1.67 | 0 | 0 | 0 |
| Fear | 4 | 0 | 84 | 8 | 0 | 4 |
| Happiness | 0 | 3.08 | 0 | 96.92 | 0 | 0 |
| Sadness | 3.33 | 0 | 0 | 0 | 96.67 | 0 |
| Surprise | 0 | 0 | 0 | 0 | 1.25 | 98.75 |

**Table 6.** Confusion matrix for facial expression recognition in percentages, using SVM with boosted features (400 AdaBoost selected features).

|  | **Anger** | **Disgust** | **Fear** | **Happiness** | **Sadness** | **Surprise** |
|---|---|---|---|---|---|---|
| Anger | 100 | 0 | 0 | 0 | 0 | 0 |
| Disgust | 1.67 | 96.67 | 1.67 | 0 | 0 | 0 |
| Fear | 0 | 0 | 92 | 4 | 0 | 4 |
| Happiness | 0 | 0 | 0 | 100 | 0 | 0 |
| Sadness | 0 | 0 | 3.33 | 0 | 96.67 | 0 |
| Surprise | 0 | 0 | 0 | 0 | 1.25 | 98.75 |

*5.4. Comparison with State-of-the-Art Methods*

The recognition accuracy achieved by the proposed method on the Cohn-Kanade facial expression dataset for facial expression recognition is comparable with the best accuracy in the literature. We achieved 95.17% of facial expression recognition accuracy using multi-class AdaBoost with DTW similarity between feature vectors based weak classifier, and 97.35% of recognition accuracy using SVM on boosted features. So far, the system in [22] has shown superior performance, and has achieved a 99.7% recognition rate. In their method, the landmark initialization is a manual process, and the number of landmarks is also larger than the number of landmarks in the proposed method. On the other hand, the proposed method is fully automatic. In [7], a 96.26% recognition rate has been achieved, using a method based on local binary patterns and SVM classifiers. The main drawback of the method in [7] is that it is only tested in perfect manually aligned image sequences, and no experiments in fully automatic conditions have been presented. Similarly in [44], a 97.16% of recognition rate has been achieved, by extracting the most discriminated facial landmarks for each facial expression. Recently, Zhao *et al.* [9] obtained 94.88% of recognition accuracy on a single facial image, using LBP features and a kernel discriminant isomap. Jabid *et al.* [14] achieved 93.69% of recognition accuracy, using local directional pattern features, which are similar to the LBP feature with SVM. Another, more recent method proposed by Zhang *et al.* [45], achieved 97.14% of recognition accuracy. The LBP features were used with sparse representation classifier. Therefore, the best recognition accuracy of the different methods proposed by researchers in the literature is around 97% (except for the method in [22]), on the Cohn-Kanade facial expression database. Our proposed method also achieved more than 97% of the recognition accuracy, which is the second best accuracy so far, at least according to the authors' knowledge.

## 6. Conclusions

Two methods for facial expression recognition, by using either multi-class AdaBoost with DTW, or by using SVM on the boosted features, are proposed in this paper. The geometric features are extracted from the sequences of facial expression images, based on tracking results of facial landmarks. The proposed facial expression recognition system is fully automatic, in which the landmark initialization and tracking is based on the EBGM method. Each extracted geometric feature vector is used to build a single weak classifier, which is based on the DTW similarity between the input feature vector and the prototypical feature vector of each facial expression. A multi-class AdaBoost algorithm is used, to select the subset of discriminative feature vectors. A recognition accuracy of 95.17% using feature selective multi-class AdaBoost, and 97.35% using SVM on boosted features, is achieved on the Extended Cohn-Kanade (CK+) facial expression database. Feature selection using AdaBoost and expression classification using SVM gives the best recognition accuracy. The discriminative feature vectors for each facial expression are also determined, and the result is analyzed, depending on the face region from which the facial landmark tracking results are contributed, to build the feature vectors.

Our experiments show that the movements of facial landmarks, as a particular expression evolves, are not independent of each other. In our system, the prototypical facial expressions are computed based on the assumption that each facial expression can be modeled using unimodel distributions. Since the facial expressions are recognized successfully, with high recognition accuracy, our assumption turns out to be true. This means that for each class of facial expression in the database, there are similar movements of landmarks, as the facial expression evolves over time, independent of the ethnic group, age, and gender.

## Acknowledgements

This work was partially supported by the National Research Foundation of Korea grant funded by the Korean government (2011-0022152).

## Conflict of Interest

The authors declare no conflict of interest.

**References**

1. Valstar, M.F.; Mehu, M.; Jiang, B.; Pantic, M.; Scherer, K. Meta-analysis of the first facial expression recognition challenge. *IEEE Trans. Syst. Man. Cybern. B Cybern.* **2012**, *42*, 966–979.
2. Mehrabian, A. Communication without words. *Psychol. Today* **1968**, *2*, 53–56.
3. Ekman, P. Strong evidence of universal in facial expressions: A reply to Russell's mistaken critique. *Psychol. Bull.* **1994**, *115*, 268–287.
4. Samal, A.; Iyenger, P.A. Automatic recognition and analysis of human faces and facial expressions: A survey. *Pattern Recognit.* **1992**, *25*, 65–77.
5. Pantic, M.; Rothkrantz, L. Automatic analysis of facial expressions: The state of the art. *IEEE Trans. Pattern Anal. Mach. Intell.* **2000**, *22*, 1424–1445.
6. Fasel, B.; Luettin, J. Automatic facial expression analysis: A survey. *Pattern Recognit.* **2003**, *36*, 259–275.
7. Zhao, G.; Pietikänen, M. Dynamic texture recognition using local binary patterns with an application to facial expressions. *IEEE Trans. Pattern Anal. Mach. Intell.* **2007**, *29*, 915–928.
8. Shan, C.; Gong, S.; McOwan, P.W. Facial expression recognition based on local binary patterns: A compressive study. *Image Vision Comput.* **2009**, *27*, 803–816.
9. Zhao, X.; Zhang, S. Facial expression recognition based on local binary patterns and kernel discriminant isomap. *Sensors* **2011**, *11*, 9573–9588.
10. Moore, S.; Bowden, R. Local binary patterns for multi-view facial expression recognition. *Comput. Vision Image Underst.* **2011**, *115*, 541–558.
11. Zhao, G.; Huang, X.; Taini, M.; Li, S.Z.; Pietikänen, M. Facial expression recognition from near-infrared videos. *Image Vision Comput.* **2011**, *11*, 607–619.
12. Ghimire, D.; Lee, J. Histogram of orientation gradient feature-based facial expression classification using bagging with extreme learning machine. *Adv. Sci. Lett.* **2012**, *17*, 156–161.
13. Dhall, A.; Asthana, A.; Goecke, R.; Gedeon, T. Emotion Recognition Using PHOG and LPQ Features. In Proceedings of the IEEE International Conference on Face and Gesture Recognition and Workshop, Santa Barbara, CA, USA, 21–25 March 2011; pp. 878–883.
14. Jabid, T.; Kabir, Md.H.; Chae, O. Robust facial expression recognition based on local directional pattern. *ETRI J.* **2010**, *32*, 784–794.
15. Zhi, R.; Flierl, M.; Ruan, Q.; Kleijn, W.B. Graph-preserving sparse nonnegative matrix factorization with application to facial expression recognition. *IEEE Trans. Syst. Man Cybernet.-Part. B Cybernet.* **2011**, *41*, 38–52.
16. Kotsia, I.; Buciu, I. Pitas, I. An analysis of facial expression recognition under partial facial image occlusion. *Image Vision Comput.* **2008**, *26*, 1052–1067.
17. Lin, D.-T. Facial expression classification using PCA and hierarchical radial basic function network. *J. Inf. Sci. Eng.* **2006**, *22*, 1033–1046.
18. Wang, Z.; Ruan, Q. Facial Expression Recognition Based Orthogonal Local Fisher Discriminant Analysis. In Proceedings of the International Conference on Signal Processing (ICSP), Beijing, China, 24–28 October 2010; pp. 1358–1361.
19. Littlewort, G.; Bartlett, M.S.; Fasel, I. Dynamics of facial expression extracted automatically from video. *Image Vision Comput.* **2006**, *24*, 615–625.

20. Choi, H.-C.; Oh, S.-Y. Realtime Facial Expression Recognition Using Active Appearance Model and Multilayer Perceptron. In Proceedings of the International Joint Conference SICE-ICASE, Busan, Korea, 18–21 October 2006; pp. 5924–5927.
21. Asthana, A.; Saragih, J.; Wagner, M.; Goecke, R. Evaluating AAM Fitting Methods for Facial Expression Recognition. In Proceedings of the International Conference on Affective Computing and Intelligent Interaction, Amsterdam, The Netherlands, 10–12 September 2009; pp. 1–8.
22. Kotsia, I.; Pitas, I. Facial expression recognition in image sequence using geometric deformation features and support vector machines. *IEEE Trans. Image Process.* **2007**, *16*, 172–187.
23. Sebe, N.; Lew, M.S.; Sun, Y.; Cohen, I.; Gevers, T.; Huang, T.S. Authentic facial expression analysis. *Image Vision Comput.* **2007**, *25*, 1856–1863.
24. Sung, J.; Kim, D. Real-time facial expression recognition using STAAM and layered GDA classifiers. *Image Vision Comput.* **2009**, *27*, 1313–1325.
25. Chang, Y.; Hu, C.; Feris, R.; Turk, M. Manifold based analysis of facial expression. *Image Vision Comput.* **2006**, *24*, 605–614.
26. Sandbach, G.; Zafeiriou, S.; Pantic, M.; Rueckert, D. Recognition of 3D facial expressions dynamics. *Image Vision Comput.* **2012**, 762–773.
27. Rudovic, O.; Pantic, M.; Patras, I. Coupled Gaussian processes for pose-invariant facial expression recognition. *IEEE Trans. Pattern Anal. Mach. Intell.* **2012**, *35*, 1357–1369.
28. Lyons, M.; Skamatsu, S. Coding Facial Expressions with Gabor Wavelets. In Proceedings of the IEEE International Conference on Automatic Face and Gesture Recognition, Nara, Japan, 14–16 April 1998; pp. 200–205.
29. Zhang, Z.; Lyons, M.; Schuster, M.; Skamatsu, S. Comparison between Geometric-Based and Gabor-Wavelets-Based Facial Expression Recognition Using Multi-Layer Perceptron. In Proceedings of the IEEE International Conference on Automatic Face and Gesture Recognition, Nara, Japan, 14–16 April 1998; pp. 454–459.
30. Huang, X.; Zhao, G.; Pietikäinen, M.; Zheng, W. Dynamic Facial Expression Recognition Using Boosted Component-Based Spatiotemporal Features and Multi-Classifier Fusion. In Proceedings of Advanced Concepts for Intelligent Vision Systems, Sydney, Australia, 13–16 December 2010; pp. 312–322.
31. Sakoe, H.; Chiba, S. Dynamic programming algorithm optimization for spoken word recognition. *IEEE Trans. Acoust. Speech Signal. Process.* **1978**, *26*, 43–49.
32. Lemire, D. Faster retrieval with a two-pass dynamic-time-warping lower bound. *Pattern Recognit.* **2009**, *42*, 2169–2180.
33. Lades, M.; Vorbüggen, J.C.; Buhmann, J.; Lange, J.; Malsburg, C.; Würtz, R.P.; Konen, W. Distortion invariant object recognition in the dynamic link architecture. *IEEE Trans. Comput.* **1993**, *42*, 300–311.
34. Wiskott, L.; Fellous, J.-M.; Krüger, N. Face recognition by elastic bunch graph matching. *IEEE Trans. Pattern Anal. Mach. Intell.* **1997**, *19*, 775–779.
35. Blome, D.S. Elastic Bunch Graph Matching. M.Sc. Thesis, Colorado State University: Fort Collins, CO, USA, 22 May 2003.
36. Viola, P.; Jones, M.J. Robust real-time face detection. *Int. J. Comput. Vision* **2004**, *57*, 137–154.

37. Freund, Y.; Schapire, R.E. A decision-theoretic generalization of on-line learning and an application to boosting. *J. Comput. Syst. Sci.* **1997**, *55*, 119–139.
38. Silapachote, P.; Karuppiah, D.R.; Hanson, A.R. Feature Selection Using AdaBoost for Face Expression Recognition. In Proceedings of the International Conference on Visualization, Image and Image Processing, Marbella, Spain, 6–8 April 2004; pp. 84–89.
39. Lajevardi, S.M.; Lech, M. Facial Expression Recognition from Image Sequence Using Optimized Feature Selection. In Proceedings of the International Conference on Image and Vision Computing, Christchurch, New Zealand, 26–28 November 2008; pp. 1–6.
40. Zhu, J.; Zou, H.; Rosset, S.; Hastie, T. Multi-class AdaBoost. *Stat Interface* **2009**, 2, 349–360.
41. Lucey, P.; Cohn, J.F.; Kanade, T.; Saragih, J.; Ambadar, Z. The Extended Cohn-Kanade Dataset (CK+): A Complete Dataset for Action Unit and Emotion-Specified Expression. In Proceedings of the 3rd IEEE Workshop on CVPR for Human Communication Behavior Analysis, San Francisco, CA, USA, 13–18 June 2010; pp. 94–101.
42. Chang, C.-C.; Lin, C.-J. *LIBSVM: A library for support vector machines*, 2001. Available online: http://www.csie.ntu.edu.tw/~cjlin/libsvm (accessed on 9 June 2013).
43. Hsu, C.-W.; Chang, C.-C.; Lin, C.-J. *A Practical Guide to Support Vector Classification; Technical Report*; Department of Computer Science, National Taiwan University, Taiwan, 2010.
44. Zafeiriou, S.; Pitas, I. Discriminant graph structures for facial expression recognition. *IEEE Trans. Multimed.* **2008**, *10*, 1528–1540.
45. Zhang, S.; Zhao, X.; Lei, B. Robust facial expression recognition via compressive sensing. *Sensors* **2012**, *12*, 3748–3761.